\documentclass[10pt]{article}
\pdfoutput=1
\usepackage[letterpaper]{geometry}
\usepackage{hicss}
\usepackage{times}
\usepackage[none]{hyphenat}
\usepackage{url}
\usepackage{latexsym}
\usepackage{indentfirst}
\usepackage{graphicx}
\usepackage{float}
\usepackage[caption = false]{subfig}
\graphicspath{{images/}}

\title{Kartta Labs: Collaborative Time Travel}

\author{Sasan Tavakkol \\
  Google Research \\
  {\underline{tavakkol@google.com}} \\\And
  Feng Han \\
  Google Research \\
  {\underline{bladehan@google.com}} \\\And 
  Brandon Mayer \\
  Google Research \\
  {\underline{bmayer@google.com}}  \\\And
  Mark Phillips \\
  Google Research \\
  {\underline{embeepea@google.com}} 
  \AND
  Cyrus Shahabi\thanks{*The work was performed when the author was a visiting researcher at Google.} \\
  Google Research \\
  {\underline{cshahabi@google.com}} \\\And
  Yao-Yi Chiang \\
  University of Southern California \\
  {\underline{yaoyic@usc.edu}} \\\And
  Raimondas Kiveris \\
  Google Research \\
  {\underline{rkiveris@google.com}} \\
}
\date{}

\begin{document}
\maketitle
\begin{abstract}
We introduce the modular and scalable design of Kartta Labs, an open source, open data, and scalable system for virtually reconstructing cities from historical maps and photos. Kartta Labs relies on crowdsourcing and artificial intelligence consisting of two major modules: Maps and 3D models. Each module, in turn, consists of sub-modules that enable the system to reconstruct a city from historical maps and photos. The result is a spatiotemporal reference that can be used to integrate various collected data (curated, sensed, or crowdsourced) for research, education, and entertainment purposes. The system empowers the users to experience collaborative time travel such that they work together to reconstruct the past and experience it on an open source and open data platform.
\end{abstract}

\section{Introduction}
The ultimate goal of Kartta Labs is to create a collaborative time travel experience; think of Google StreetView (or Google Earth), but with the ability to go far back in time~\cite{ethington2007placing}. As with StreetView, our system needs to run on top of a map service; however, any map service we use must support a temporal dimension. Therefore the first step in this project is building a modular and scalable system to collect, process, and serve map data indexed by time and space. The Maps project consists of a stack of web applications that crowdsources collecting~\cite{goodchild2007citizens} and vectorizing historical maps. The vectorized spatiotemporal data are open sourced to promote the collaboration among the community. These vectorized data are also served online using a tile server\footnote{https://www.ogc.org/standards/wmts} and visualized within a map renderer website. We previously introduced some parts of the Maps module in \cite{tavakkol2019kartta}.

The second step in this project is to reconstruct the historical buildings as 3D models. To this end, we introduced an image processing pipeline in \cite{kapoor2019nostalgin} where the first step was an image segmentation job to identify buildings facades. The identified facades were then fed to rectification~\cite{zhang2013single} and inpainting~\cite{yu2019free} jobs. The output image was then applied on a face of a cuboid 3D mesh as a texture. In this paper, we introduce our improved pipeline which extracts 3D features of the facades and incorporates accurate footprints from historical maps. Our pipeline segments and parses a single view image of the building to procedurally reconstruct a 3D mesh of its facade. Subsequently, this facade is incorporated into one face of a 3D extrusion of the building footprint. The result is stored as a 3D mesh model in an online repository, accessible through a public API.

We follow the principles of systems design to layout the architecture of Kartta Labs and build a modular system. The modules and their sub-modules are primarily defined based on their input and output. The output of one sub-module becomes the input to another, creating an organic workflow. We also outline the storage and processing requirements of each module and briefly discuss their implementation. As we mentioned earlier, our system consists of two major modules: Maps and 3D models. Each module consists of smaller sub-modules. Figure~\ref{fig:warper} shows the major modules, their sub-modules, and the workflow. We briefly explain each sub-module in this paper. Most of the sub-modules are open sourced and as they mature, are added to our GitHub organization (https://github.com/kartta-labs). We use Kubernetes to deploy and manage our tools, which makes it easy for others to redeploy our suite of applications either for development or production purposes. We currently run an experimental instance of our system on Google Cloud Platform and are planning to launch a full version by the end of 2020.

\begin{figure}[t]
    \centering
	\includegraphics[width=1\linewidth]{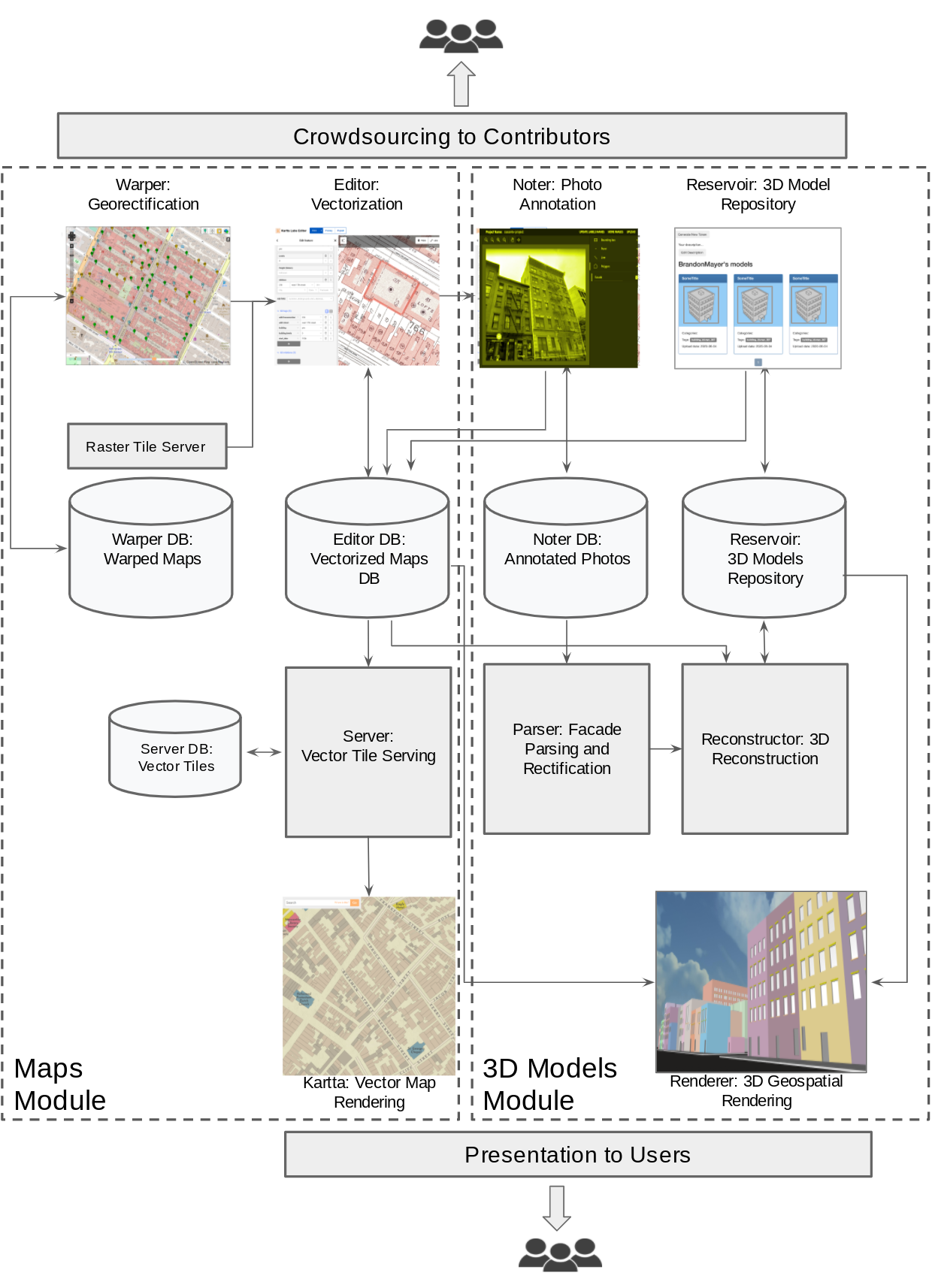}
	\caption{Sub-modules of the Kartta Labs software system.}
	\label{fig:overall}       
\end{figure}

\section{Related Work}

A geographical backdrop (e.g., maps) to be used as a reference for integration of other datasests has always been of interest to researchers. This is evident by the numerous mashups developed on top of Google Maps.
As a natural extension to this spatial reference, some use cases consider a dynamic spatiotemporal reference system. 
For example, Gapminder (www.gapminder.org), has a map feature that allows the user to geospatially visualize a statistical value (e.g., population) and navigate it through time using a time-slider feature. Another example is HistoryPin, a crowdsourced archive of historical media. HistoryPin enables users to "pin" their media (e.g., images) to Google Maps and set the time to which they belong. Kartta Labs can act as a platform for such systems, providing accurate historical geospatial data over time as reference.

Endeavors to construct 3D worlds have been pursued for decades in academia and industry. Virtual worlds are examples of such endeavors that became relatively popular in 1990's and are recently gaining traction again, thanks to the advances in virtual reality devices. Examples of virtual worlds are Active Worlds, Second Life, World of Warcraft among others~\cite{sivan20083d3c}. The geography of these virtual worlds are often a fantasy world. For example, Active Worlds consists of hundreds of fantasy worlds, where users can explore 3D virtual environments built by others or build their own 3D content. Active Worlds has inspired several academic efforts in education~\cite{holmstrom2001using, dickey2005three} and has served as a platform for data collection for various studies~\cite{naper2001system}. In contrast, Kartta Labs is meant to reconstruct the real world in time and in space.

Esri's CityEngine\footnote{https://en.wikipedia.org/wiki/CityEngine} is another related work to Kartta Labs. CityEngine takes a procedural approach to construct 3D urban environments. It can procedurally generate 3D models given footprints of buildings. While the generated 3D models look compelling and consume metadata such as buildings height, they are not based on real world imagery and therefore the building facades are not detailed. CityEngine does not natively support a time dimension or tiling. Indeed, applications like CityEngine can be used to generate 3D models for Kartta Labs.

Another closely related work to Kartta Labs is 3DCityDB~\cite{yao20183dcitydb}, a free 3D geo-database solution for 3D city models based on CityGML standard issued by the Open Geospatial Consortium (OGC). 3DCityDB does not natively support a historical time dimension. As we discuss in Section \ref{conclusionAndFutureWork} we are considering using 3DCityDB to host city 3D models of Kartta Labs in the future.

Google Earth is perhaps the closest application to what we envision. Google Earth renders a 3D representation of Earth primarily based on contemporary satellite imagery. In addition to represent the surface of earth in 3D, Google Earth, also shows 3D building models in some cities. At the beginning, a community was formed around Google Earth that used applications such as SketchUp and Building Maker to manually create the 3D buildings, resembling our crowdsourcing approach to the problem. However, it now uses auto-generated 3D models. Google Earth also enables users to explore historical satellite imagery going back a few decades. However, it does not represent the historical satellite imagery in 3D, nor does vectorize them.

To the best of our knowledge, Kartta Labs is the only system that is capable of not only vectorizing historical maps, but also reconstructing them in 3D across time. Most of the similar solutions are focused on contemporary data. Others either deal with only maps or 3D reconstruction~\cite{kapoor2019nostalgin}. Furthermore the most compelling solutions are based on proprietary code and data. Kartta Labs differentiates itself from the prior work by combining the features of several similar applications and providing them as an open source and open data platform. 

\section{Design}
We designed Kartta Labs following the principles of systems design to create a modular and scalable\footnote{Scalability is the ability of a system to handle more work by adding more resources} software system. A modular design was required for Kartta Labs for several reasons. First, Kartta Labs mission is quite complicated. Therefore, as any software, a modular design let us divide the problem to smaller pieces and solve them independently. More importantly, our modular design enables us to adopt open source solutions for some of the modules. Furthermore, having a well defined interface between modules let us have more than one implementation for a module. For example, our photo annotation module has two implementations, one based on crowdsourcing and one based on artificial intelligence. Finally, a modular design makes Kartta Labs scalable.

We define our system and its modules based on their inputs and outputs, enabling us to define clean interfaces between modules. The input to Kartta Labs, as a system, is historical photos and maps. The output is a 3D representation of world with a time dimension. In order to process the input and create the output, Kartta Labs may rely on intermediate inputs such as geotagging and georefrencing of the input images and maps by humans.

Kartta Labs consists of two major modules: Maps and 3D models. In Section \ref{maps_section} we describe the Maps module and its sub-modules. The input of this module is a scanned historical map and the output is the same map, but in vector format. In Section \ref{3d_section} we layout the architecture of our 3D models module. The vector historical maps generated by the Maps module becomes the input to 3D models module. Furthermore, the 3D models module takes in historical urban photos as its input. The output of this module is the overall output of Kartta Labs: a 3D representation of world with a time dimension. We briefly explain the sub-modules of Maps and 3D models in their corresponding sections.

Karrta Labs is implemented in several different languages using different technologies and development frameworks. This is because we leveraged available open source solutions that are developed within different communities and perhaps for unrelated purposes. However, we unified the deployment of all these applications using Docker containers\footnote{https://www.docker.com/resources/what-container} and Kubernetes\footnote{https://kubernetes.io/}. This deployment design not only makes our system a portable solution, such that it can be deployed locally or on different cloud platforms (e.g., Google Cloud), but also enables it to scale out\footnote{Scaling out or horizontal scaling is adding more nodes (e.g., virtual machines.) to a system to handle more work} and scale up\footnote{Scaling up or vertical scaling is adding more resources to a single node by, for example, increasing its number of CPU's, memory, or disk storage.} on demand.

We use Google Cloud Platform (GCP) to deploy Kartta Labs. In addition to its Kubernetes Engine we use GCP's managed databases and storage to leverage its scalability, security, and reliability. We also use Google Clouds Functions, a serverless execution environment for running simple, single-purpose cloud services, for some of our simple services. Nevertheless, Kartta Labs can be deployed on other cloud platforms or locally on a single machine for development purposes.

\section{Maps}
\label{maps_section}
The Maps module aims to create a map server with a time dimension, we envision OpenStreetMap\footnote{https://openstreetmap.org} (OSM) with a time slider to navigate the time dimension. We have developed and stacked a set of open source tools that are used to collect and vectorize scanned historical maps, via crowdsourcing, and serve them as vector tiles\footnote{https://en.wikipedia.org/wiki/Vector\_tiles}. Maps is made up of a suite of tools that allow users to upload historical maps, georectify them to match real world coordinates, and then convert them to vector format by tracing their geographic features. These vectorized maps are then served on a tile server and rendered as maps in the browser.

The input of the Maps module is a scanned historical map and the output is the same map, but in vector format. The entry point of the Maps module is a web application, called Warper, that enables the users to upload historical images of maps and georectify them by finding control points on the historical map and corresponding points on a base map. Another web application, Editor, allows users to load the georectified historical maps generated by Warper as the background (through a raster tile server) and then trace their geographic features (e.g., building footprints, roads, etc.). These traced data are stored in OSM vector format. They are then converted to vector tiles and served from a vector tile server, dubbed as Server. Finally, our browser map renderer, called Kartta, visualizes the spatiotemporal vector tiles allowing the users to navigate space and time on historical maps. We briefly discuss the design of each Maps sub-module next.

\subsection{Georectification}
\begin{figure}[t]
    \centering
	\includegraphics[width=1\linewidth]{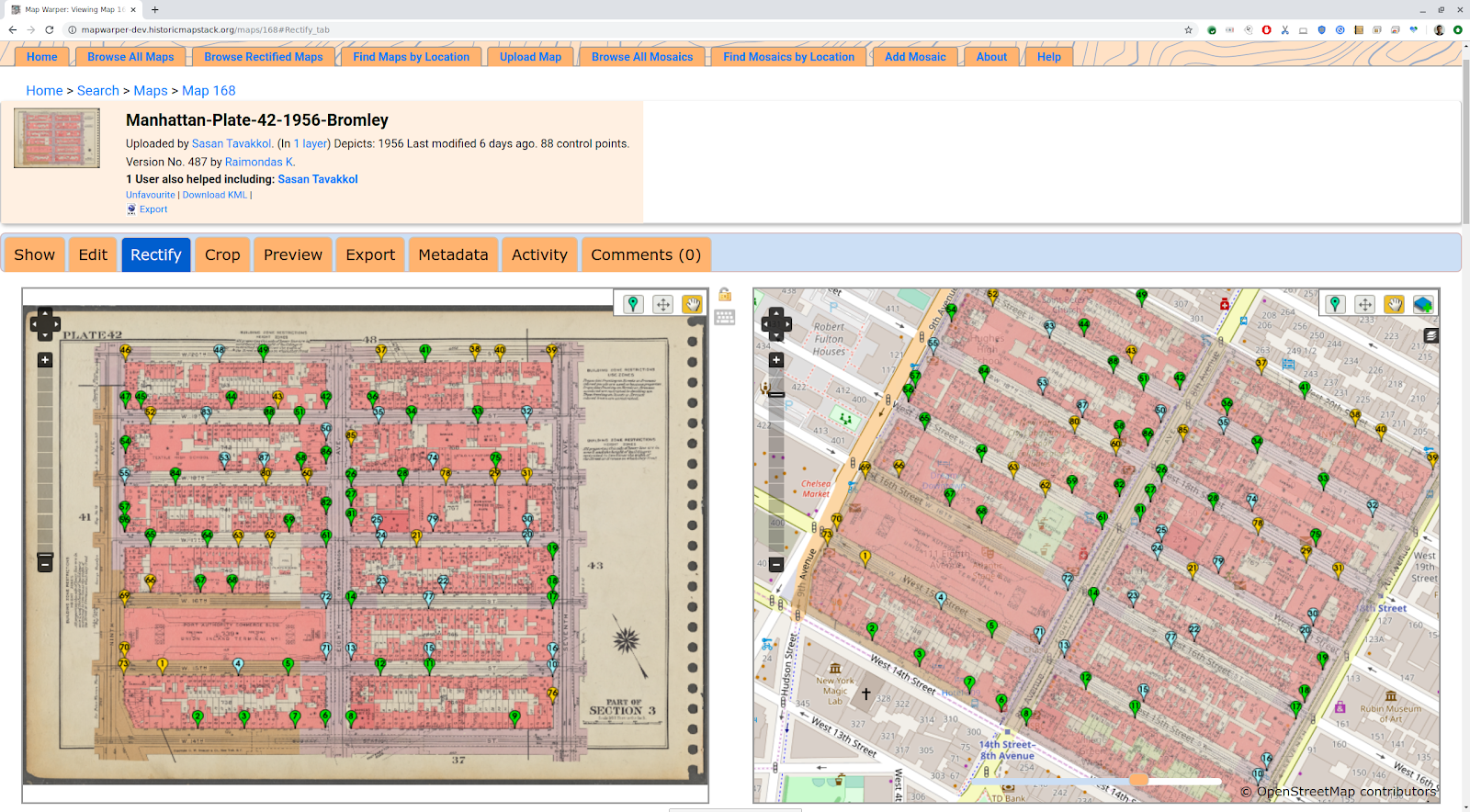}
	\caption{A screenshot of Warper showing how a map is georectified.}
	\label{fig:warper}       
\end{figure}

Warper is an open source web application that crowdsources collection and georectification of historical maps. It is based on the MapWarper\footnote{https://github.com/timwaters/mapwarper} open source application. The input of Warper is scanned historical maps that users may upload. Warper makes a best guess of an uploaded map's geolocation by extracting textual information from the map and using algorithms outlined in \cite{tavakkol2019kartta, zekun2020an}. This initial guess is used to place the map roughly in its location and let the user georeference the map pixels by placing pairs of control points on the historical map and a reference map.\footnote{Note that advanced 2D georectification processes, such as~\cite{Chen2004-gq}, can be used to automatically identify control point pairs and can be later added to Warper.} Given the georeferenced points, the application warps the image such that it aligns well with the reference map. This georectified map is the output of this sub-module. Warper also runs a raster tiles server that serves each georectified map at a tile URL. This raster tile server is used to load the georectified map as a background in the Editor application that is described next. Figure~\ref{fig:warper} shows a screenshot of Warper where a historical map of New York is georeferenced against a contemporary map of the same area from OSM. 


\subsection{Vectorization}
\begin{figure}[t]
    \centering
	\includegraphics[width=1\linewidth]{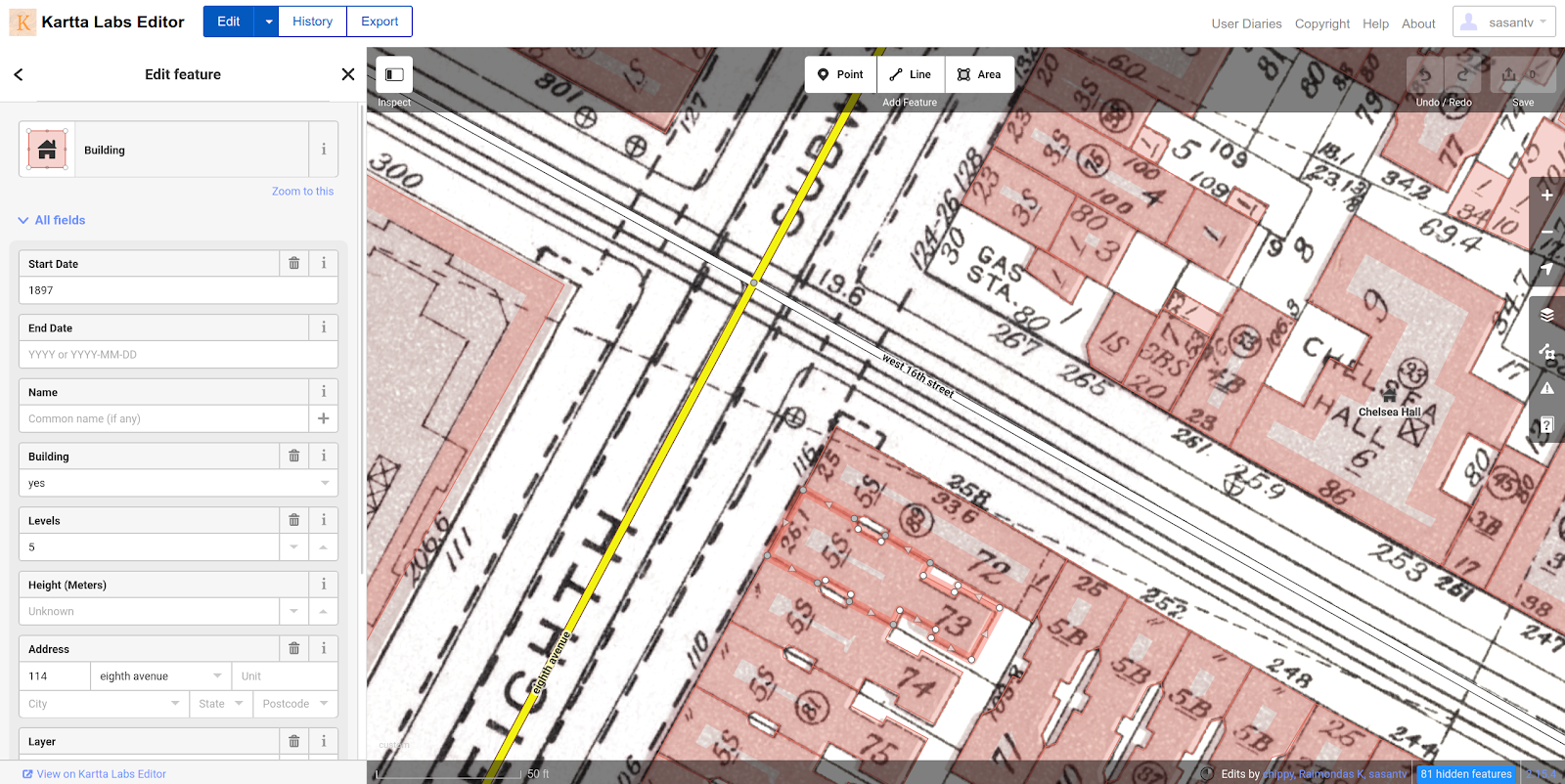}
	\caption{A screenshot of the Editor application used to trace footprints on a historical map.}
	\label{fig:editor}       
\end{figure}

The Editor is an open source web application from OSM\footnote{https://github.com/openstreetmap} stack of tools that we have modified to fit in our system. Editor lets users extract vector geometries (output) from georectified images (input) and then stores them in a database. The vector data include information such as buildings footprints, roads, addresses, names and dates, as well as "start date" and "end date" fields which represent the time dimension; a feature is considered to exist in time between these two dates.
A screenshot of the Editor web application is shown in Figure~\ref{fig:editor}.

\subsection{Tiling}
To support the development of interactive map applications with a time dimension, we serve our spatiotemporal map data (input) as a collection of Mapbox vector tiles\footnote{https://docs.mapbox.com/vector-tiles/reference/} (output) using the Tegola\footnote{https://tegola.io/} vector tile server. We call this application Server, for short. This service makes tiles available using the standard OSM tile naming convention\footnote{https://wiki.openstreetmap.org/wiki/Slippy\textunderscore map\textunderscore tilenames}.

In our current implementation the time dimension is included as an attribute on the tile data; tiles are addressed by space (and zoom level) only
Client applications can present a view of the data for a specific moment in time by using the "start date" and "end date" attributes to filter out features not present at that moment.
 
\subsection{Visualization}
\begin{figure}[t]
    \subfloat[1930]{\includegraphics[width = 1.5in]{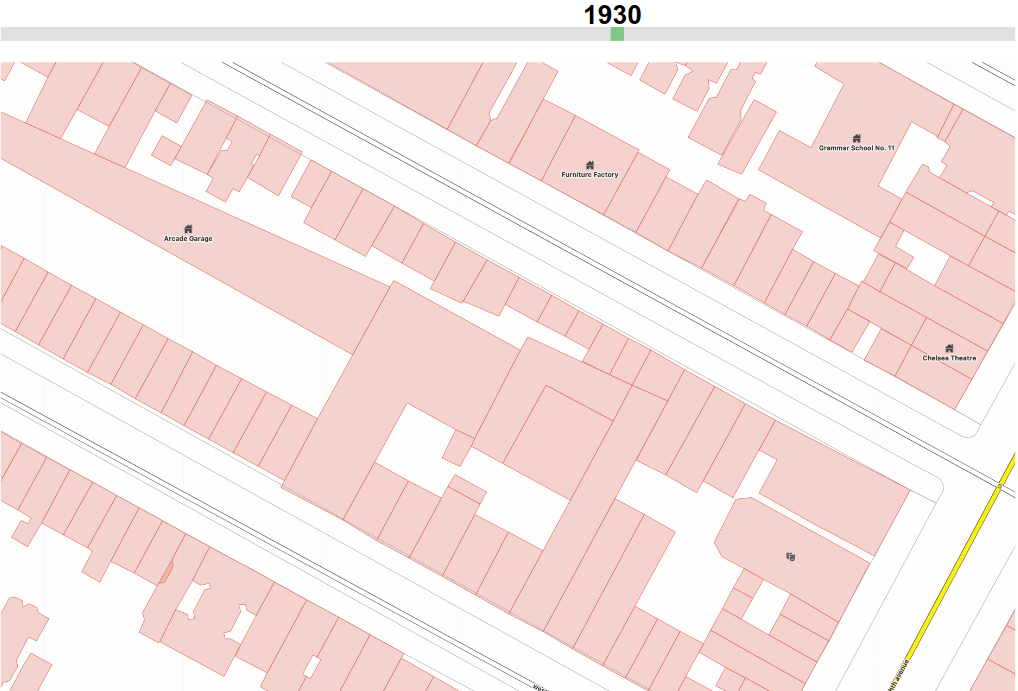}} 
    \subfloat[1932]{\includegraphics[width = 1.5in]{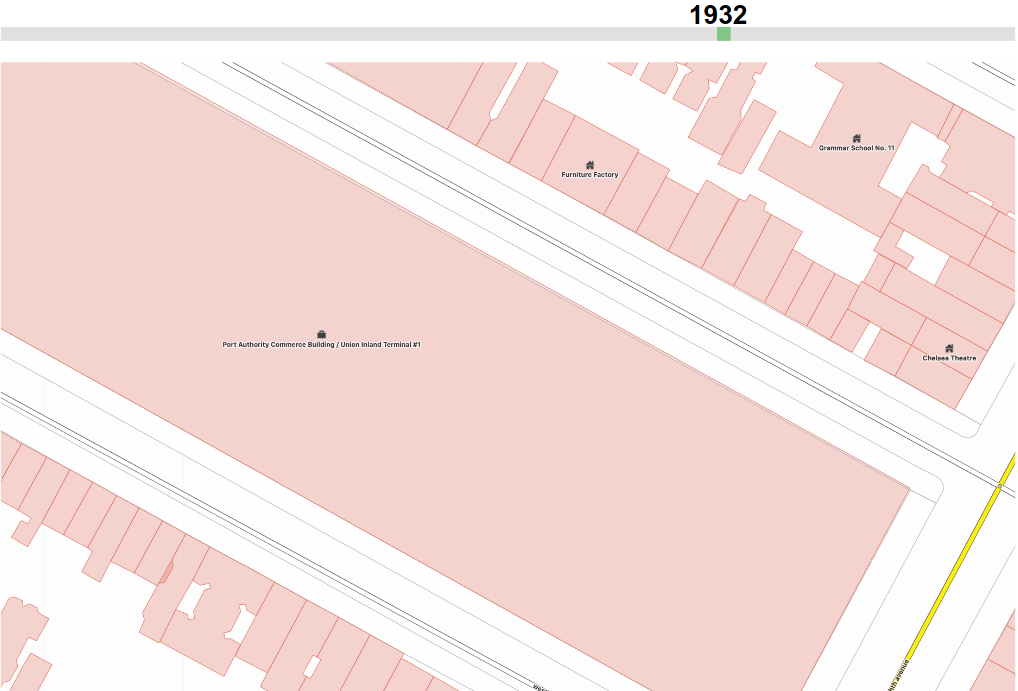}}\\
    \caption{Screenshots of the Kartta showing the area of the Google NYC building in Manhattan, before (a) and after (b) it was built.}
    \label{fig:viz}
\end{figure}

The endpoint of the Maps module is a time-aware, interactive map application, called Kartta. Kartta works like any familiar map application (e.g., Google Maps), but also has a time slider so the user can choose the time at which they want to see the data. By moving the time slider, the user is able to see how features in the map such as buildings and roads have changed over time. The input to Kartta is a set of vector tiles and the output is rendered images showing those tiles in a given map style. Note that the images are rendered client-side, i.e., in the browser. Figure~\ref{fig:viz} shows two snapshots of this application in two different times around the Google NYC building (111 Eighth Avenue, New York, NY). Generating vector tiles, as opposed to raster tiles, was required to provide a seamless navigation of the time dimension with any granularity.

\section{3D Models}
\label{3d_section}
The 3D Models module aims to reconstruct the detailed full 3D structures of historical buildings using the associated images and maps data, organize these 3D models properly in an online repository, and render them on the historical maps with a time dimension. The input to this module is historical images and vector historical maps, and the output is a 3D representation of an area across time.

\begin{figure}[t]
    \centering
	\includegraphics[width=1\linewidth]{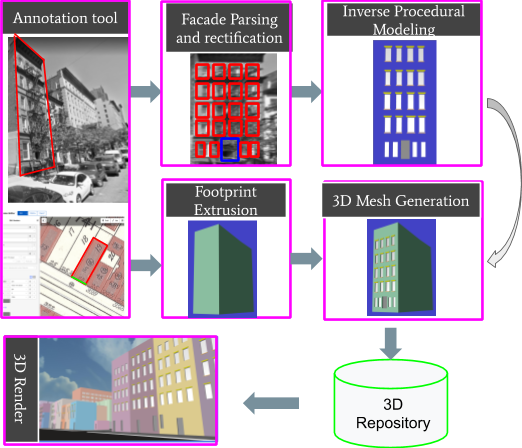}
	\caption{System diagram for the 3D models module.}
	\label{fig:sys}       
\end{figure}

In most cases, there is at most one historical image available for a building, which makes the 3D reconstruction an extremely challenging problem. To tackle this challenge, we developed a “coarse-to-fine reconstruction-by-recognition” algorithm as illustrated in Figure~\ref{fig:sys}. The footprint of the building is extruded upwards to generate the coarse 3D structure, using any available metadata (e.g., number of floors) to set the extrusion height. Then, the historical image is annotated, either by crowdsourcing or automated algorithms, and the result is used to generate 3D details (e.g. windows, entrances, stairs) for the visible facades of the building from the street ground level. 
We discuss each sub-module next.

\subsection{Photo Annotation}
We need to annotate the historical photos to identify building facades and then to identify the structural details of each facade. We rely on crowdsourcing and machine learning algorithms. To crowdsource the annotation task, we developed a web application, called Noter. It consists of a frontend based on the open source tool MakeSense \footnote{https://github.com/SkalskiP/make-sense} connected to a backend we developed in Python. The application allows users to upload photos of historical buildings (input) or browse the photos uploaded by others. Users can then annotate (output) the photos given a preset of labels (facade, window, door, etc.). 
An ID is assigned to each annotation piece such as facades. The facade ID is used to associate that facade with part of a footprint in the Editor application. This process geotags that specific facade but it can also be used to roughly geolocate the rest of the facades in the same photo. If a facade is matched with another one in a different photo as being same, then the location information can be propagated between those photos. We are working on a spatial reasoning algorithm to construct a graph of facades and propagate the location information from one facade to others in the same sub-graph~\cite{tavakkol2020Piaget}. Such an algorithm can significantly facilitate geotagging historical photos.

\subsection{Facade Parsing and Rectification}
Facade parsing is the process of decomposing a facade into its constituent components, e.g., windows, entries, doors, and stair. We call our facade parsing sub-module Parser. The input to this sub-module is the photo of a building facade and the output is a rectified photo of the same facade with its components fully annotated. We take a supervised learning approach. A corpus of approximately 5,000 images were annotated by human annotators with over 500,000 boundary-level instance annotations.

We trained binary FasterRCNN neural networks using the facade component annotations for each target semantic class which are used to localize bounding-box level instances in new images. We used binary FasterRCNN rather than a single multi-class detector due to our observations of superior performance of a suite of binary classifiers compared to the multiclass version on held out data.

\begin{figure*}[t]
    \begin{center}
    \includegraphics[width=\linewidth]{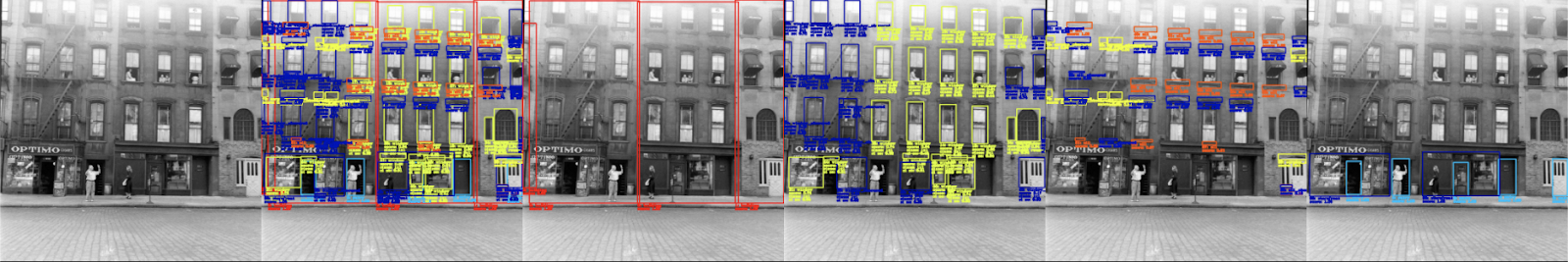}
    \end{center}
	\caption{Kartta Labs' facade parsing output. The input image (far left) is parsed to detect facade sub-components such as windows, window sills, cornices, roof cornice, storefronts, entries, and stairs.}
	\label{fig:facade-parsing}       
\end{figure*}

While extremely accurate, the FasterRCNN model is only capable of producing axis aligned (relative to the image frame) bounding box localizations requiring a rectification post-processing step. We have had success training and integrating semantic segmentation models including DeepLab~\cite{chen2017deeplab} into the Kartta Labs Facade parsing pipeline but defer discussions of semantic segmentation for later publications. Figure~\ref{fig:facade-parsing} visualizes the output of the facade parsing pipeline prior to rectification and 3D reconstruction. The facade Parsing pipeline is written in C++ using the open-source MediaPipe\footnote{https://github.com/google/mediapipe} framework. The MediaPipe framework allows for parallelization and thread optimization of image processing routines on a per-process basis. 

After parsing an image into facade components, the next step in the pipeline is to extract each facade primitive within the target (annotated) facade and normalize them with respect to camera viewpoint. We use a vanishing-point based rectification process to bring all components within each facade into frontal view. 
Man-made objects like facades have strong regularities and follow architectural principles. We use predefined grammar rules to regularize the parsing results on the rectified facade. 
For example, we organize windows in a grid and force them to share the same dimensions and appearance (e.g. number of panels, cornices, and sills), across each row.

\subsection{3D Reconstruction}
As illustrated in Figure~\ref{fig:sys}, the 3D reconstruction sub-module, dubbed as Reconstructor, consists of the following parts:

\noindent\textbf{Footprint extrusion}: The inputs to this part is a footprint polygon and its metadata (e.g. number of floors). We convert the geocoordinates of the footprint into Mercator coordinates and then into meters. We extrude the footprint vertically considering the height of the building to output the its coarse 3D mesh.

\noindent\textbf{Inverse procedural modeling}: The inputs to this part is the parsed sub-components (e.g. windows, entries, stairs, etc.) within a rectified facade. For each sub-component category, we first extract a set of parameters (e.g. width/height ratio for windows) and then use procedural modeling to generate a 3D instance of this category to provide a realistic 3D experience consistent with the given image.

\noindent\textbf{3D mesh generation}: With the help of the annotation tool, each annotated facade is also linked to one side of the footprint and thus linked to a 3D plane of the footprint extrusion. With this correspondence, we can compute a transformation that maps a point on the rectified facade to its corresponding point on the face of the footprint extrusion. Using this transformation, we map each reconstructed 3D sub-component to the proper location on the footprint extrusion. At the end, we can merge these transformed 3D sub-component and footprint extrusion into one single mesh as the final 3D reconstruction of the target building.

\subsection{3D Model Repository}
The Kartta Labs’ 3D Model Repository, called Reservoir, hosts and serves the geolocated 3D models for downstream rendering. It is an open-sourced web service, based on the 3DMR\footnote{https://gitlab.com/n42k/3dmr} project, that hosts the reconstructed 3D assets which can be inspected, modified, pushed, and fetched either through a user interface or programmatically through a REST API. An ID is associated with each 3D model uploaded to the Reservoir which can be used to link it to a building footprint in Editor. Unlike other sub-modules in Kartta Labs, Reservoir does not process its input (3D models) to generate an output.

Reservoir is a centralized location for federated researchers to push their temporal and geolocated reconstructions with corresponding metadata to a common platform for uniform downstream rendering. This decoupling extends to the rendering process as the open-sourced 3D assets served by the model repository can be accessed and rendered by multiple, potentially independent rendering projects. 


\subsection{3D Rendering}
The 3D renderer of Kartta Labs, called Renderer for short, is our user facing web application that visualizes the reconstructed 3D models on their geolocation. Renderer is a client-side application that fetches the map features, including building footprints, from our database. It then extrudes a footprint if a 3D model is not available for that building, otherwise it downloads the associated 3D model from the Reservoir and renders it. The input to Renderer is the vector map tiles and the 3D models, and the output is 3D visualization of an area. Renderer uses THREE JS library to display the 3D models. To provide a fast and seamless transition in time, Renderer downloads the 3D models for all the buildings disregarding their start and end dates. It then deactivates the buildings not present in a given time set by a slider. First-person street level view and bird’s-eye view are available.

\subsection{Data}

\textbf{Collecting:} Data plays a major role in this project. Even though we rely on our users to collect historical data, we are actively looking for available resources for historical maps and urban photos. To bootstrap our historical maps database, we are discussing possible collaborations with archives, libraries, municipalities, etc. to load their archived maps and photos into our pipeline. Furthermore, some parts of the contemporary OSM data are relevant. For example, most of the streets in large cities have not changed in the past decades or there are many century-old buildings in Manhattan, New York. This kind of data is readily available in the OSM database.

\textbf{Quality control:} Quality control often becomes a critical issue in any crowdsourcing project. Furthermore, any data generated using machine learning approaches also needs proper quality control as these methods are inherently meant not to be perfect. Since Kartta Labs uses both crowdsourcing and machine learning to generate its output data, it needs to have a procedure for quality control.

Quality is a subjective issue in general. The expectations for different aspects of quality such reliability, accuracy, relevancy, completeness, and consistency~\cite{allahbakhsh2013quality} can significantly vary for different projects. For example Kartta Labs tolerates incomplete data with the expectation that it will eventually achieve completeness. As an another example, we do not need to precisely know the dates the historical photos are taken. This is because buildings life often spans several decades and it is usually enough to know the approximate snapped time of a historical photo to associate it with a set of certain buildings.

Similar to projects such as OpenStreetMap and Wikipedia, the quality control in Kartta Labs heavily relies on crowdsourcing itself. For example, users can leave "notes" on the map to describe discrepancy or correct the flawed data themselves. We also rely on automated tools to ensure the quality of our output. For example, the Editor has a feature to detect overlapping buildings. We are extending this feature to take the time dimension into account. The result is that the editing user receives a warning if a building overlaps another one at the same time period. Another example is our regularization sub-module that applies a certain set of predefined rules to ensure the reconstructed facades follow expected grammars.

Several crowdsourcing projects rely on reputation~\cite{kazemi2013geotrucrowd} of users to ensure the quality of their work. We took a similar but simpler approach by defining pipelines to ban users with malicious activity and making a small subset of users as admins with more authority. We intent to expand our quality control after we launch and collect more data.

\textbf{License:} To encourage the collaborative nature of our project, we use the Open Database License (ODbL) on our Maps data. Other generated and crowdsourced data, such as 3D reconstructions and photo annotations are also open sourced under appropriate licenses.

\section{Results} 
\begin{figure}[t]
    \centering
    \subfloat[1910]{\includegraphics[width=0.25\textwidth]{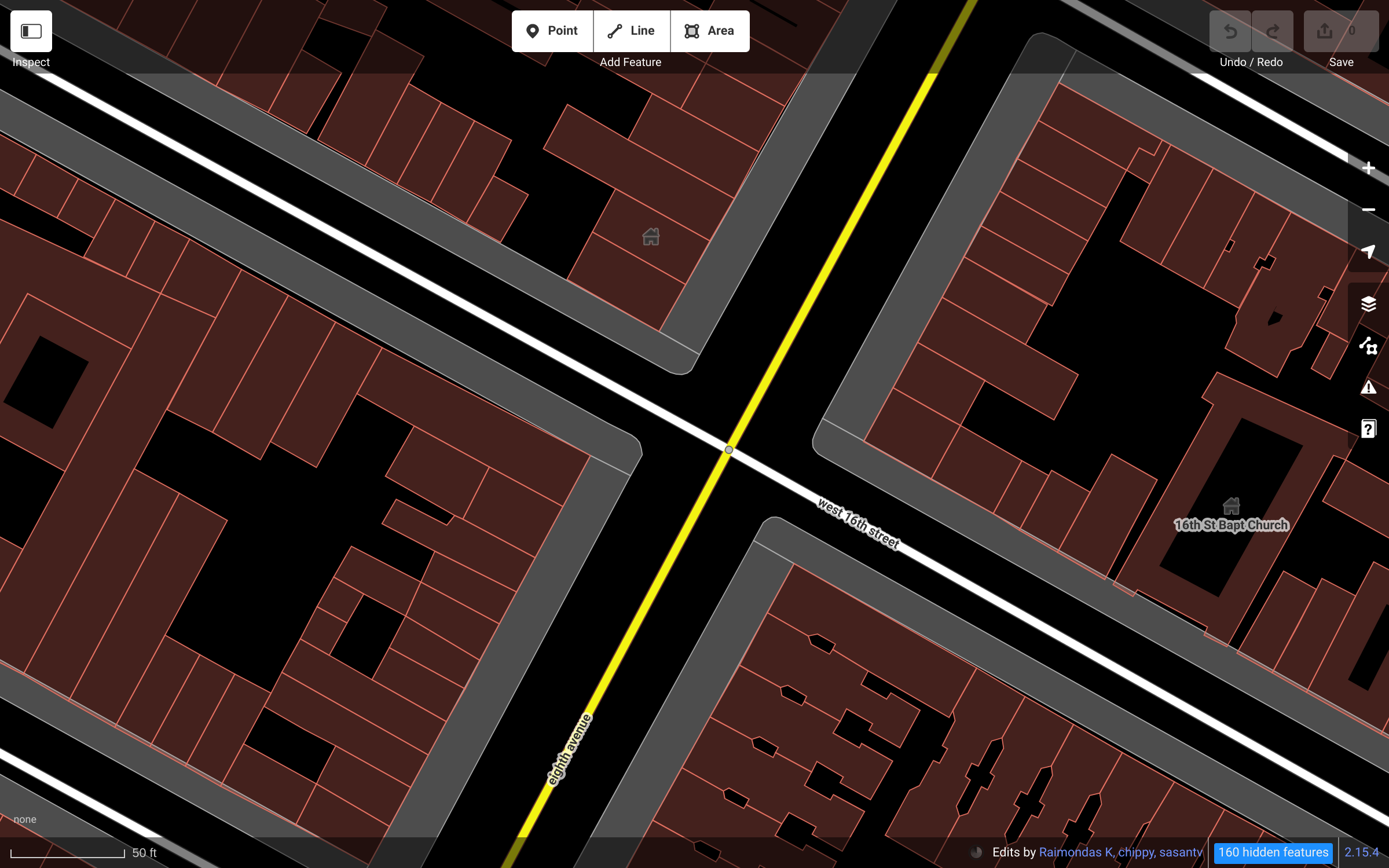}}
    \subfloat[1920]{\includegraphics[width=0.25\textwidth]{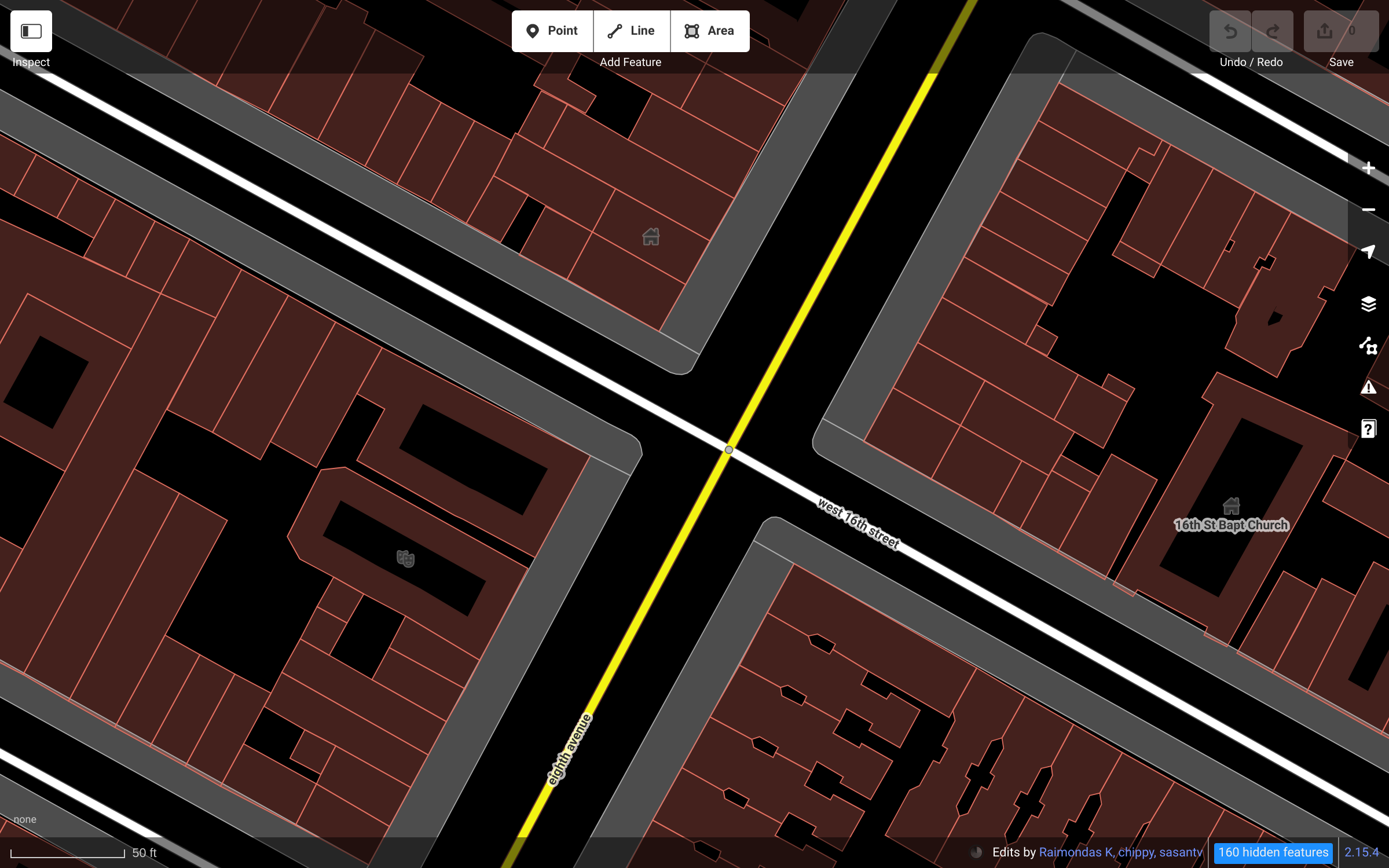}} \quad
    \subfloat[1930]{\includegraphics[width=0.25\textwidth]{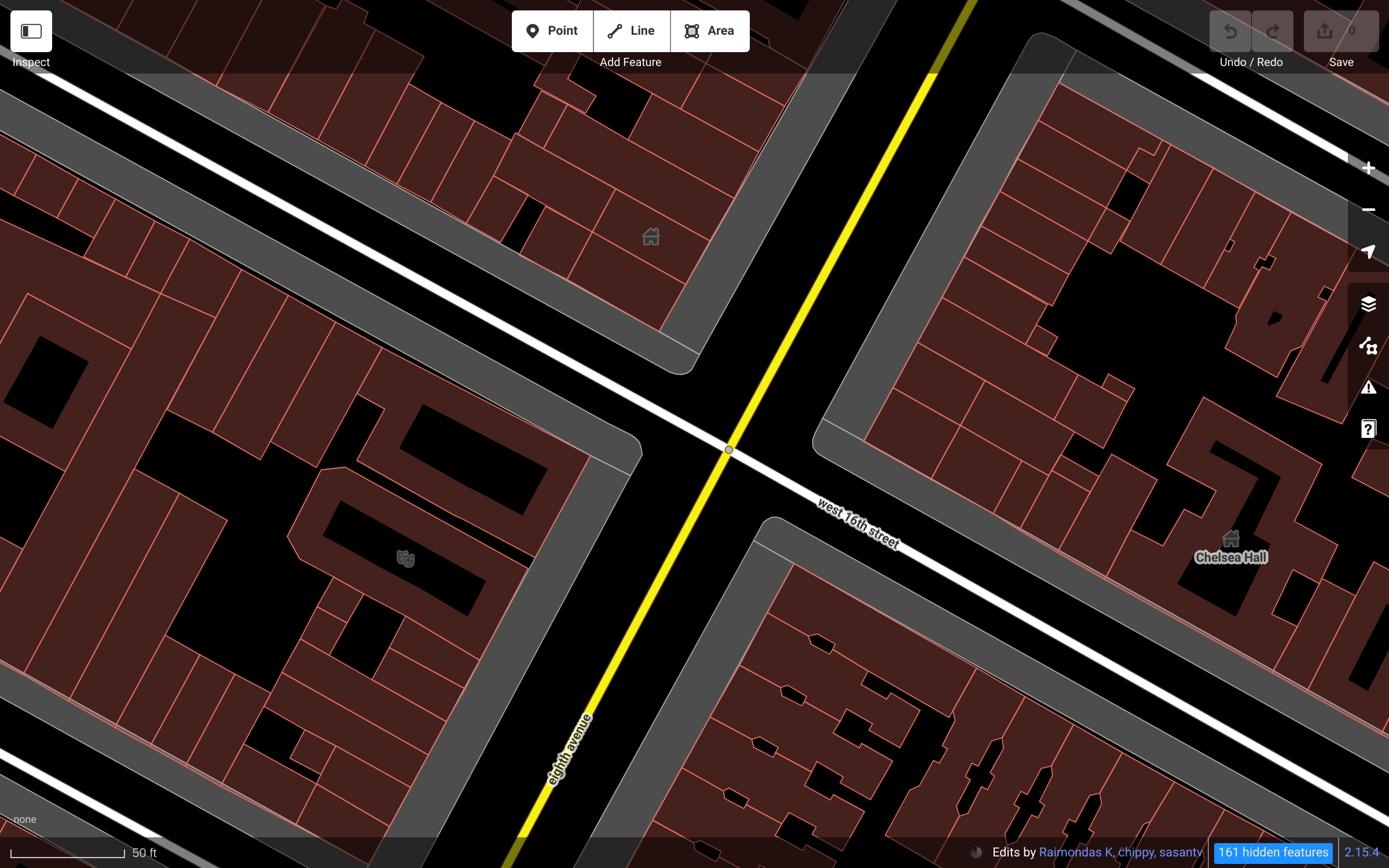}}
    \subfloat[1940]{\includegraphics[width=0.25\textwidth]{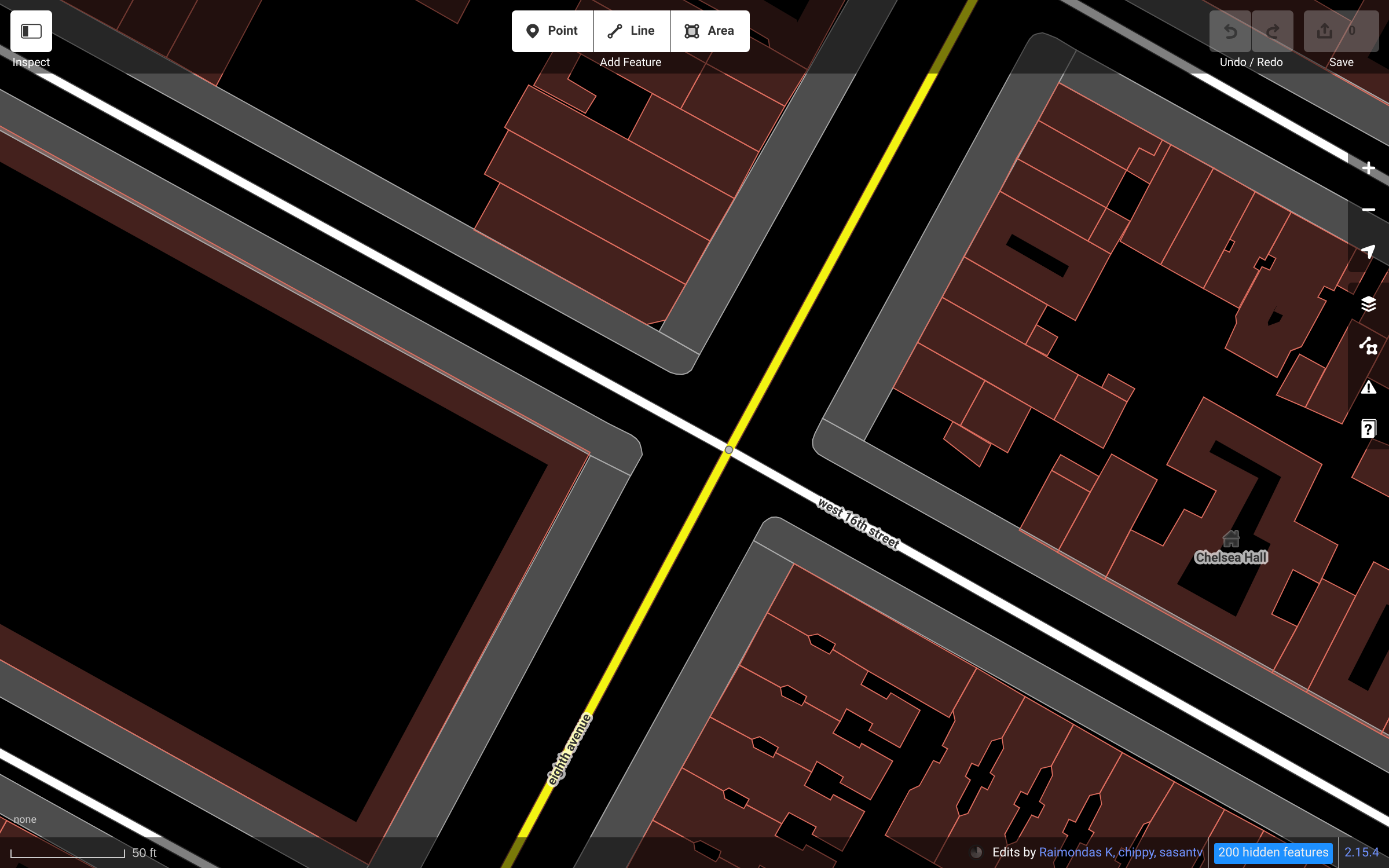}}
    \caption{Vectorized maps of part of Manhattan around Google NYC building in different years.}
	\label{fig:results-map}       
\end{figure}

\begin{figure}[t]
    \centering
    \subfloat[1910]{\includegraphics[width=0.25\textwidth]{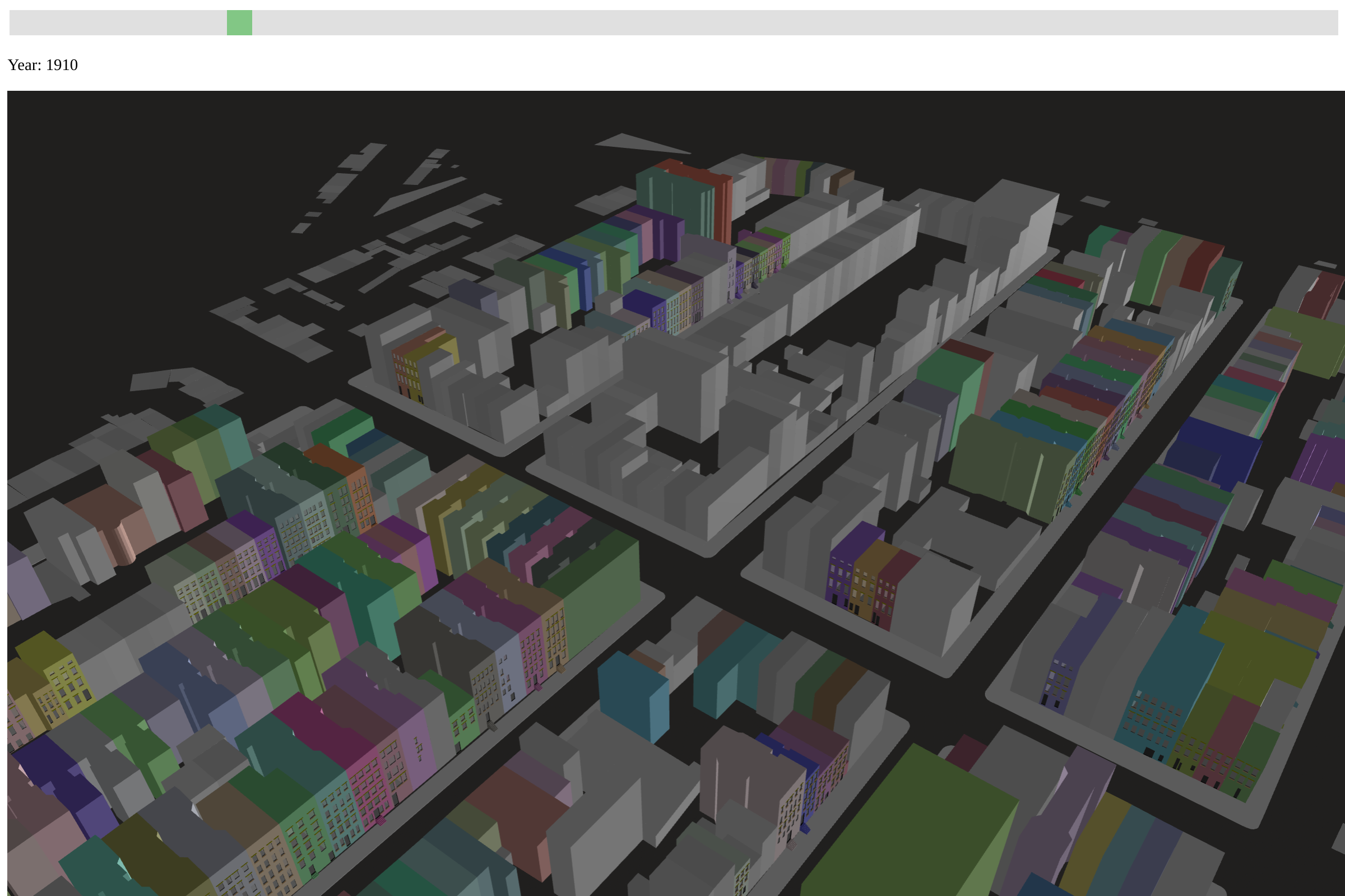}} 
    \subfloat[1920]{\includegraphics[width=0.25\textwidth]{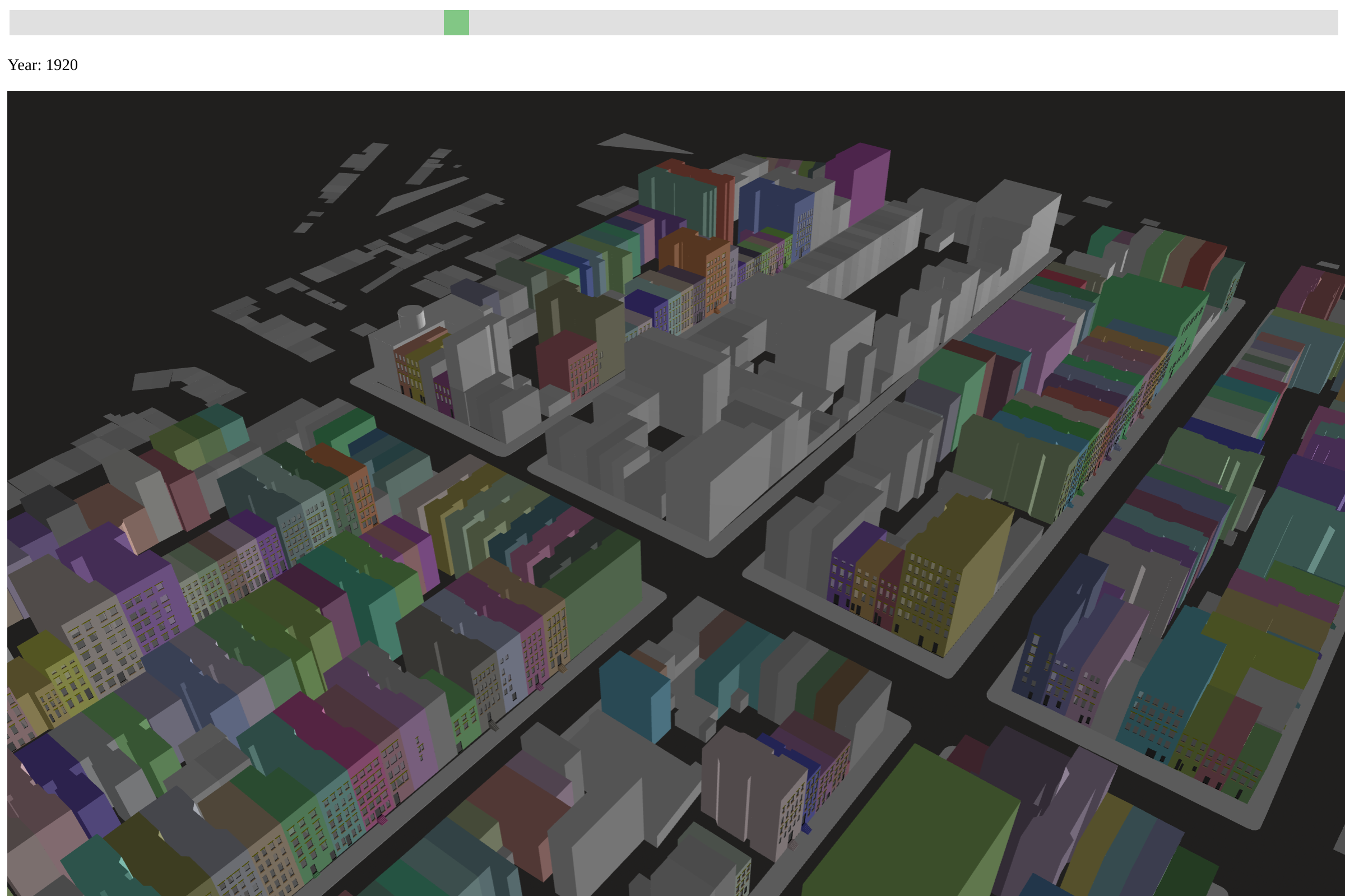}} \quad
    \subfloat[1930]{\includegraphics[width=0.25\textwidth]{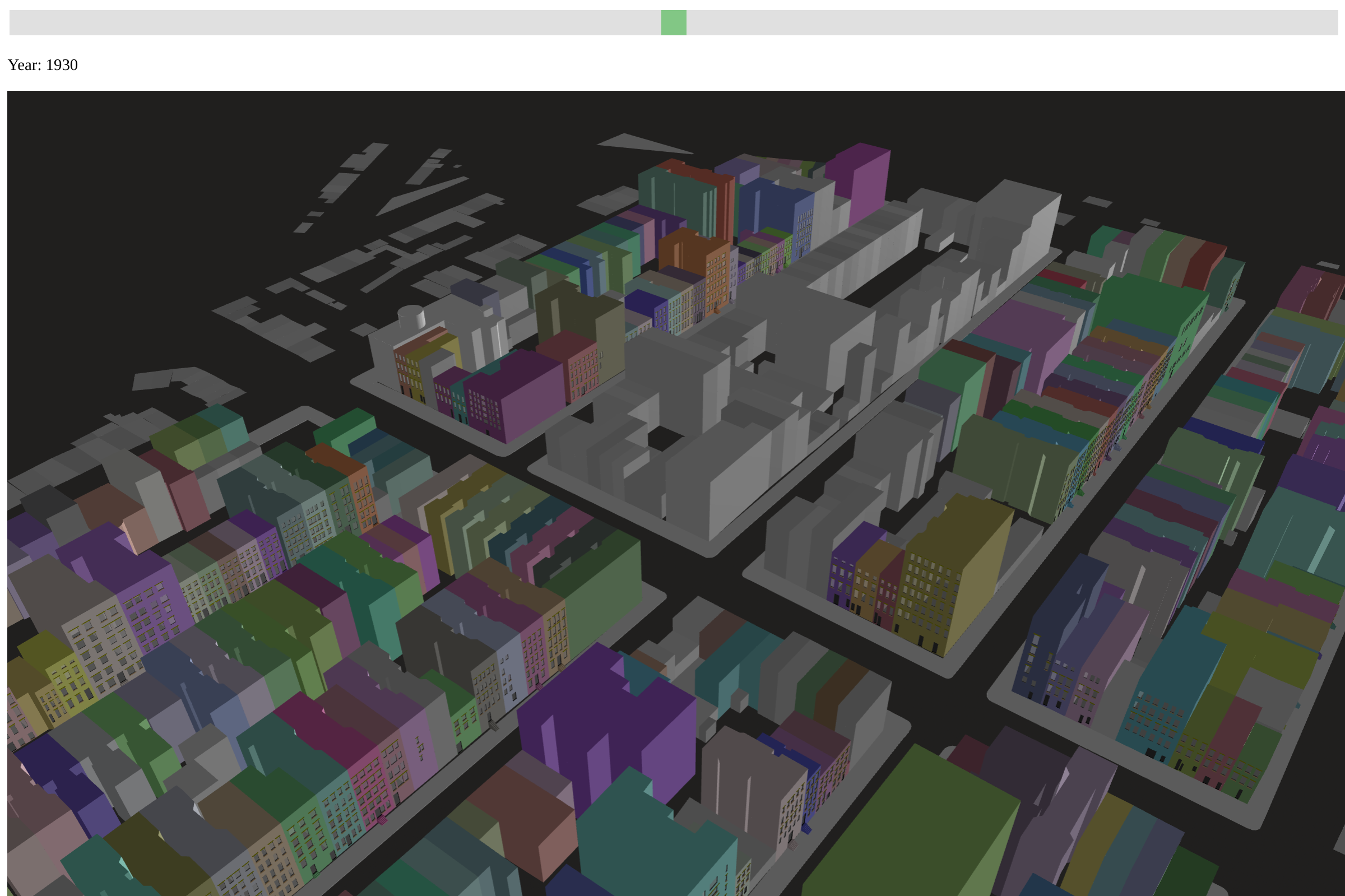}} 
    \subfloat[1940]{\includegraphics[width=0.25\textwidth]{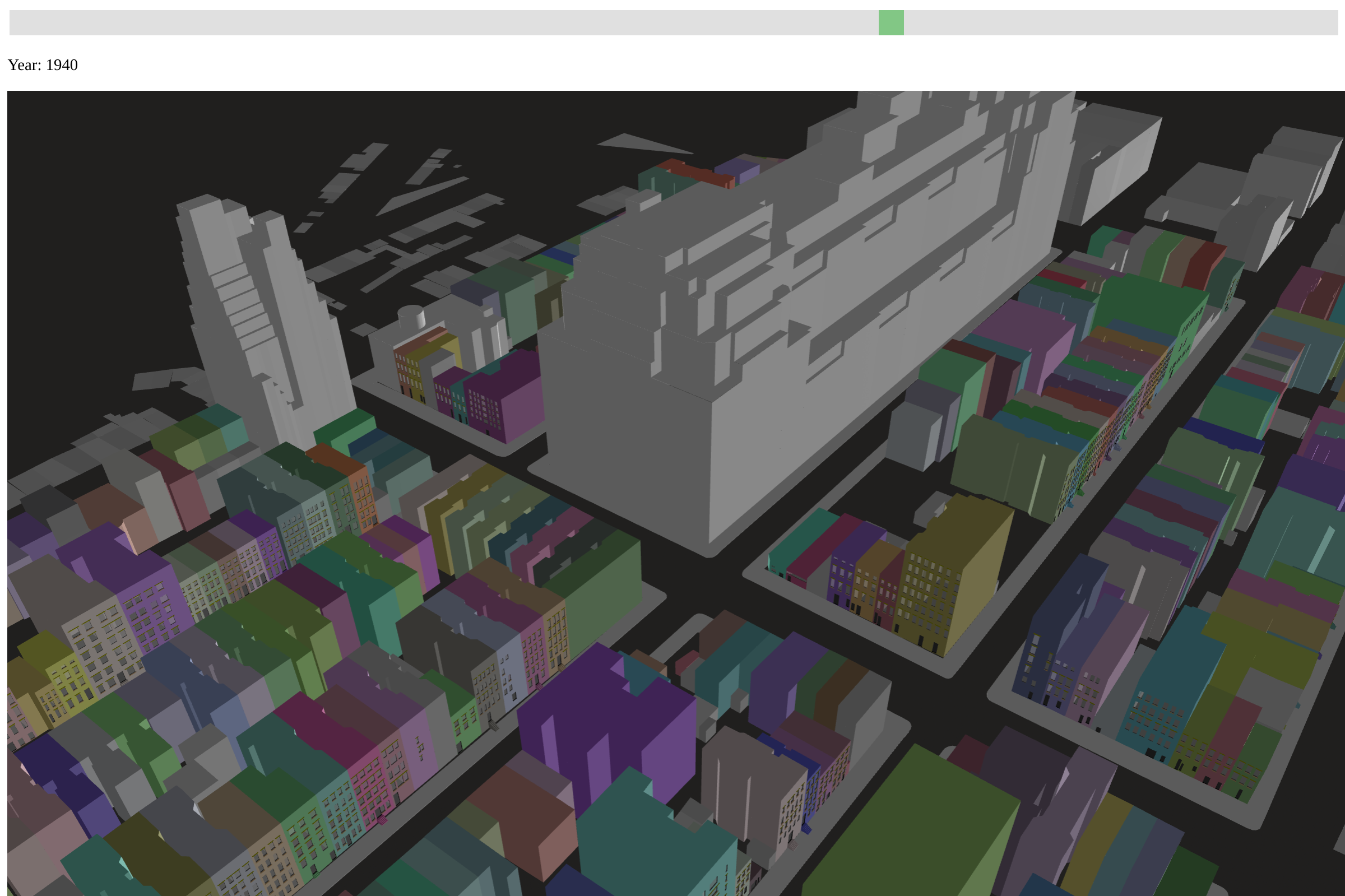}} 
    \caption{Part of Manhattan around Google NYC building reconstructed in 3D in different years from birds-eyeview of Renderer.}
	\label{fig:results-birds}       
\end{figure}

\begin{figure}[t]
    \centering
    \subfloat[1910]{\includegraphics[width=0.25\textwidth]{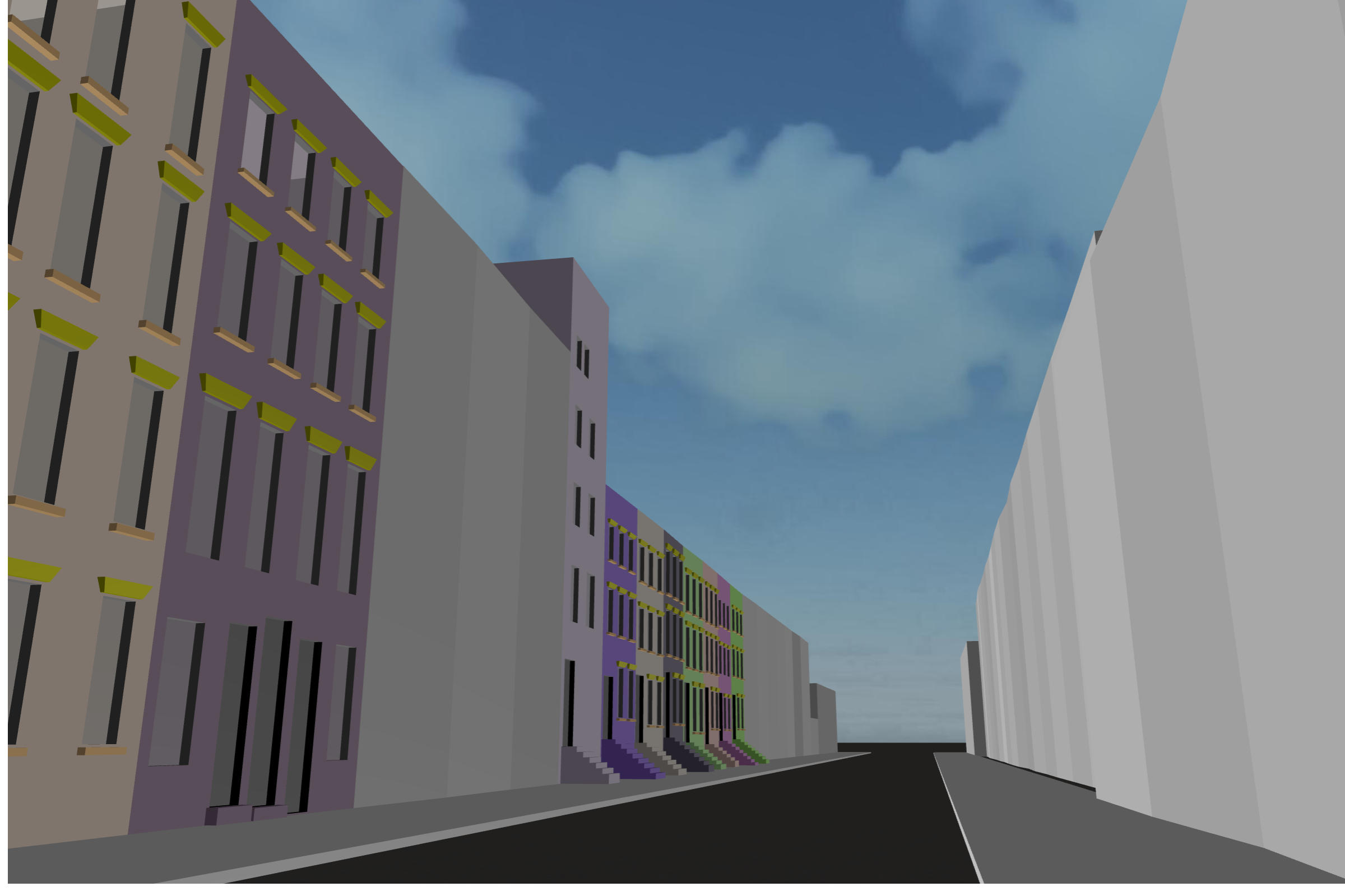}} 
    \subfloat[1920]{\includegraphics[width=0.25\textwidth]{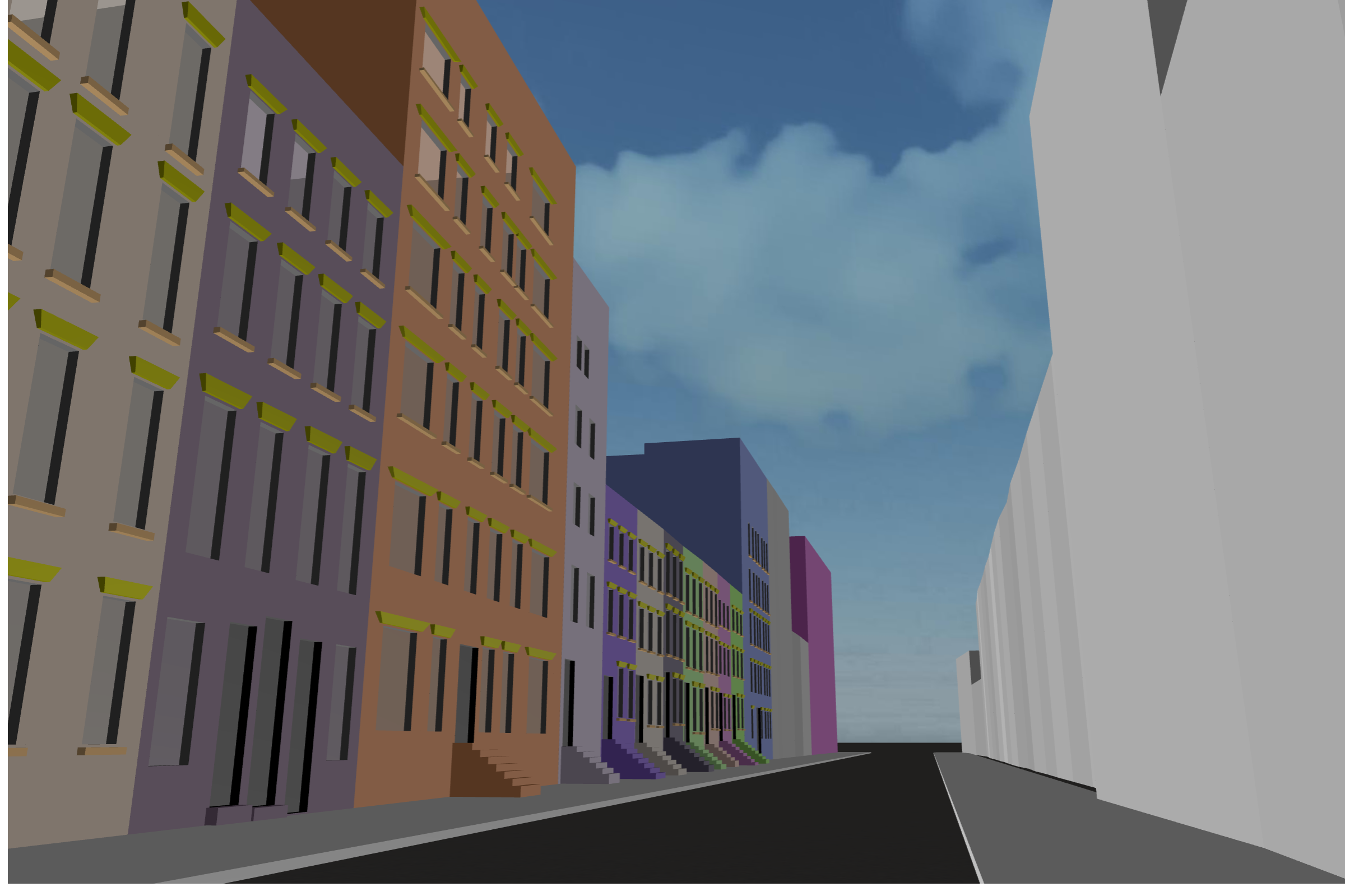}} \quad
    \subfloat[1940]{\includegraphics[width=0.25\textwidth]{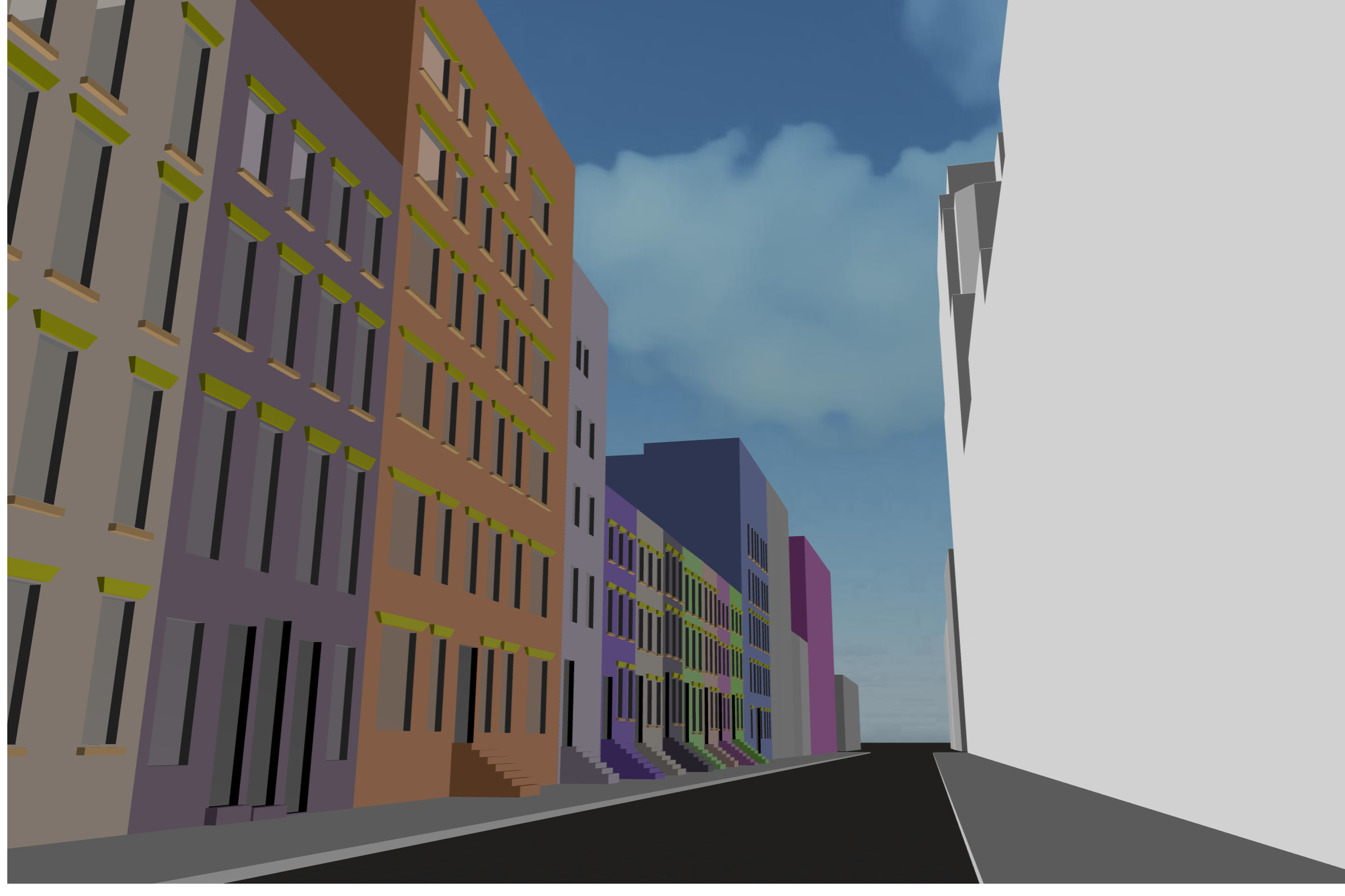}}
    \subfloat[2019]{\includegraphics[width=0.25\textwidth]{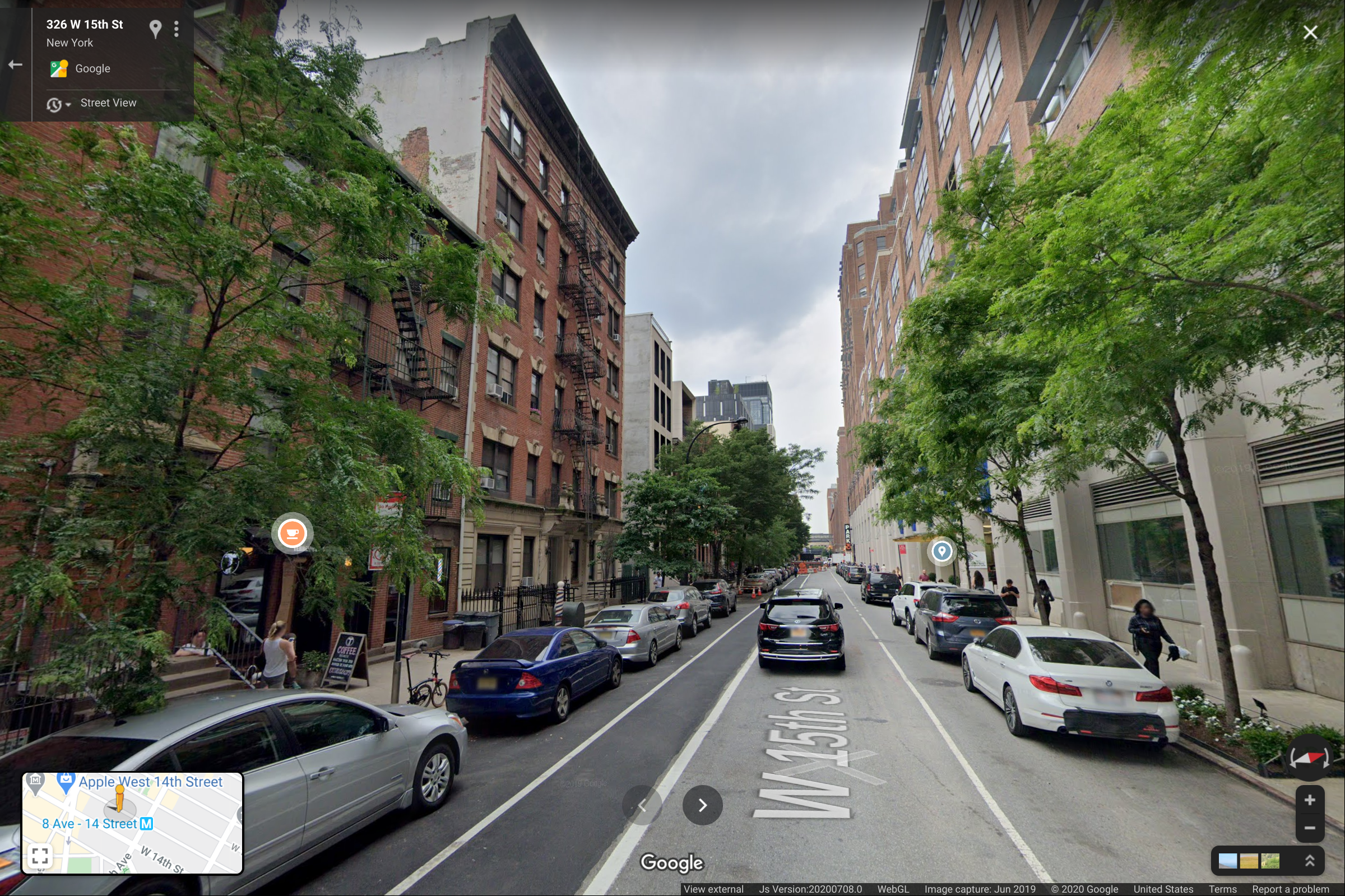}} 
    \caption{Part of Manhattan south to the Google NYC building reconstructed in 3D in different years from street level view of Renderer (a,b,c) and the same area in 2019 from Google StreetView.}
	\label{fig:results-street}       
\end{figure}

To evaluate our system, we are running an experimental instance of the Kartta Labs applications on an internal network. We reconstructed 8 blocks of Manhattan around the Google NYC building. More specifically we reconstructed the blocks between 7th and 9th avenues and W. 14th and W. 18th streets. The time was limited between 1900 to 1960. More than 1000 building footprints were traced from historical maps of different years. We were able to reconstruct the 3D models of 333 buildings from historical photos. Figure~\ref{fig:results-map} shows the map of the area north-east of the Google NYC building (intersection of 8th Ave and W. 16th Street) in 1910, 1920, 1930, and 1940. The vectorized data are extracted from scans of historical maps. Figure~\ref{fig:results-birds} shows an area around the Google NYC building during the same years but in 3D and from Renderer. We have added man-made and more accurate 3D models for a couple of buildings, including the Google NYC building, to Reservoir as a reference as well as to show the capability of the system to incorporate external 3D models. Finally, Figure~\ref{fig:results-street} shows a reconstructed street view of the 15th street, south of Google NYC building, in 1910, 1920, and 1940 from our Renderer and compares it with the modern Google StreetView of the same location. Reconstructed buildings from photos are shown in vivid colors to distinguish them from those that are only extrusions of footprints. Note that our results shown in this paper are considered preliminary. We are working on rendering our results in a photorealistic mode by generating textures for buildings facades and sub-components.

\section{Proposed Use Cases and Applications} 
The Kartta Labs system provides a valuable platform and resource for research and education. First and foremost, we would like to build a community that not only utilizes our historical datasets and open-source codes, but also contributes to both. 
As a platform that collect, integrate, and visualize detailed historical information about places, Kartta Labs can be used to facilitate numerous educational and research applications and use cases, such as topics in sociology (e.g., \cite{Kurashige2013-jv}), cancer and environmental epidemiology (e.g., \cite{Mack2004-mh}), urbanization, biodiversity (e.g., \cite{Hill2009-ee}), human disease (e.g., \cite{Yoshida2014-wt}), and biology (e.g., \cite{Davis2015-gb,Lavoie2013-ob,Vellend2013-ro}). (See~\cite{chiang2020using} and~\cite{Gregory2007-pw} for examples on using historical geographic datasets and historical Geographic Information System in scientific studies.)

We consider Kartta Labs as the underlying frame of reference to integrate various sources of spatiotemporal historical data such as traffic~\cite{DBLP:conf/iclr/LiYS018}, census, weather, crime~\cite{chainey2013gis}, pollution~\cite{perez2012near} and other environmental, sensed~\cite{DBLP:conf/gis/HeBAABCM18}, or crowdsourced~\cite{DBLP:journals/vldb/TongZZCS20} data with location and time dimensions. Imagine Kartta Labs as a generalization of Google Maps where instead of showing the current state of affairs (e.g., current traffic, current population), can show the same information for past historical time frames. For example, transportation authorities can study the impact of building certain freeways in Los Angeles on its traffic or pollution. This spatial integration of data to its historically relevant underlying infrastructure (buildings and roads) can revolutionize the way we do research and educate~\cite{10.1145/2756547}. Beyond its educational and research applications it can be used for journalism~\cite{DBLP:journals/geoinformatica/WeiSS20} and entertainment to tell better and more visually accurate stories. 

Kartta Labs can be used for change detection~\cite{hussain2013change} in various application domains from urban planning to transportation and public health~\cite{basara2008community} policy making. The decision makers can visualize seamlessly how the urban structure has changed over time and study the impact of these changes on the city infrastructure and public. For example, how often and in which locations new hospitals were built, the rate of increase (or decrease) in parks, schools, shops and restaurants in certain neighborhoods.

Finally, entertainment can be a major use case of Kartta Labs. For example, location-based games such as Ingress can extend their maps in the time-dimension, augmented reality games such as Minecraft Earth can pull in historical 3D buildings, etc. Movie industry can use Kartta Labs to recreate accurate and photorealistic historical scenes. 

\section{Conclusion and Future Work} \label{conclusionAndFutureWork}
In this paper we introduced Kartta Labs, an open source platform to reconstruct historical cities in 3D. In order to make the system open source, we designed Kartta Labs in a modular way with clear interface design (e.g., input and output) for each module, so that each module can be developed independently, potentially by extending existing open-source components, or be replaced easily in future by alternative implementations and designs. Moreover, by deploying each module in a Docker container managed by Kubernetes, we empowered Kartta Labs to both scale out and up with the ability to be deployed locally on a single machine or on different cloud platforms (e.g., Google Cloud). We also described the two main modules of the system: Maps and 3D Models. The main challenge in developing these modules is the lack of sufficient historical data, especially historical photographs from which 3D models of historical buildings can be constructed. Therefore, we are relying on an active community that can contribute data (and code) and help with geotagging historical buildings and georectifying historical maps. We developed several tools to facilitate these crowdsourced activities. The final outcome has the potential to revolutionize how we teach history and geography, how we research urban planning, transportation, and public health and how we tell stories in journalism and for entertainment. 

We are working on developing a better database schema to share our 3D models. Currently our 3D models are hosted individually on an online repository. This is useful as it enables users to view and possibly edit individual 3D models. However, it is not the most efficient solution when it comes to rendering these 3D models on a map. We are considering 3D tiling technologies such as 3DCityDB \cite{yao20183dcitydb}.

We intend to develop a number of new tools to help with automatic geotagging of historical buildings. This is challenging as the facade of the historical buildings may have changed over time and hence image-matching approaches such as PlaNet~\cite{weyand2016planet} cannot work on this dataset. The ultimate goal is to allow users to upload any historical photograph of buildings and automatically use the facade of the buildings in the picture to improve the 3D models at the correct time frame. We are also interested in expanding the community around Kartta Labs and supporting new applications and use-cases.

\section{Acknowledgements} 
We thank Amol J. Kapoor for his contributions to this project and for his thorough review of this paper.


\bibliographystyle{ieeetr}
\bibliography{references}

\end{document}